\documentclass[letterpaper, 10 pt, conference]{ieeeconf}  

\IEEEoverridecommandlockouts                              

\usepackage[nocompress]{cite}

\usepackage{amsmath,amssymb,amsfonts}
\usepackage{algorithmic}
\usepackage{graphicx}
\usepackage{textcomp}
\usepackage{xcolor}
\usepackage{multirow}
\usepackage{booktabs}
\usepackage{subcaption}
\usepackage[utf8]{inputenc}
\usepackage{float}
\usepackage{xurl} 
\usepackage[hidelinks]{hyperref} 
\usepackage{tikz}
\usepackage{tikzpagenodes}

\newcommand{\IEEECopyrightText}{%
  \footnotesize
  © 2026 IEEE. Personal use of this material is permitted.  Permission from IEEE must be obtained for all other uses, in any current or future media, including reprinting/republishing this material for advertising or promotional purposes, creating new collective works, for resale or redistribution to servers or lists, or reuse of any copyrighted component of this work in other works.
}

\AddToHook{shipout/firstpage}{%
  \begin{tikzpicture}[remember picture, overlay]
    \node[
      anchor=north,
      text width=0.94\paperwidth,
      align=center,
      font=\footnotesize
    ] at ([yshift=-7mm]current page.north) {\IEEECopyrightText};
  \end{tikzpicture}
}
\title{\LARGE \bf
GAPG: Geometry Aware Push-Grasping Synergy for Goal-Oriented Manipulation in Clutter
}

\author{Lijingze Xiao$^{\dagger}$, Jinhong Du$^{\dagger}$, Yang Cong$^{*}$, Supeng Diao, Yu Ren
\thanks{$^{\dagger}$indicates equal contributions.}
\thanks{$^{*}$The corresponding author is Prof. Yang Cong. The work was supported in part by the National Key R\&D Program of China under Grant 2023YFB4704800, NSFC under Grants 62225310, and was supported by Guangdong S\&T Program(2025B1111130001).}
\thanks{Lijingze Xiao, Jinhong Du, Yang Cong, Supeng Diao, Yu Ren are with the College of Automation Science and Engineering, South China University of Technology, Guangzhou, China. (email:\{\nolinkurl{xl24232221}, \nolinkurl {djh020503}, \nolinkurl {congyang81}, \nolinkurl{diaosupeng}, \nolinkurl{renyu0414}\}\nolinkurl{@gmail.com})}%
}
\def\BibTeX{{\rm B\kern-.05em{\sc i\kern-.025em b}\kern-.08em
    T\kern-.1667em\lower.7ex\hbox{E}\kern-.125emX}}

\begin{document}

\maketitle
\thispagestyle{empty}
\pagestyle{empty}

\begin{abstract}
Grasping target objects is a fundamental skill for robotic manipulation, but in cluttered environments with stacked or occluded objects, a single-step grasp is often insufficient. To address this, previous work has introduced pushing as an auxiliary action to create graspable space. However, these methods often struggle with both stability and efficiency because they neglect the scene’s geometric information, which is essential for evaluating grasp robustness and ensuring that pushing actions are  safe and effective.  To this end, we propose a geometry-aware push–grasp synergy framework that leverages point cloud data to integrate grasp and push evaluation. Specifically, the grasp evaluation module analyzes the geometric relationship between the gripper’s point cloud and the points enclosed within its closing region to determine grasp feasibility and stability. Guided by this, the push evaluation module predicts how pushing actions influence future graspable space, enabling the robot to select actions that reliably transform non-graspable states into graspable ones. By jointly reasoning about geometry in both grasping and pushing, our framework achieves safer, more efficient, and more reliable manipulation in cluttered settings. Our method is extensively tested in simulation and real-world environments in various scenarios. Experimental results demonstrate that our model generalizes well to real-world scenes and unseen objects. The code and video are available at https://github.com/xiaolijz/GAPG.
\end{abstract}

\section{Introduction}
Grasping is one of the most fundamental and common tasks in robotic manipulation, widely applied in various daily scenarios. However, in cluttered environments with limited grasping space, low object visibility, and diverse object poses and spatial layouts, achieving effective grasping with a single action is often challenging. To alleviate the spatial constraints in such environments, researchers have introduced auxiliary actions, such as pushing, as a collaborative strategy to actively adjust the spatial structure around the target object and improve grasp feasibility. A common approach is to use pushing to alter the position or orientation of an object, thereby creating more available grasping space~\cite{b1,b16,b17,b18,b19,b10,b29}. Currently, most methods utilize heightmap as input and employ reinforcement learning to learn grasping strategies driven by experience~\cite{b2, b20,b21}. These methods typically use the Q-value of grasp actions to provide reward signals for training auxiliary actions like pushing~\cite{b20, b21, b22}. However, some of these methods~\cite{b1,b2,b30} due to the lack of geometric awareness in grasping models, their stability in cluttered environments is limited, and their Q-values tend to be noisy and unreliable, restricting the learning process of auxiliary actions and reducing the overall task execution efficiency and success rate.
\begin{figure}[t!]
    \centering
    \includegraphics[width=0.48\textwidth]{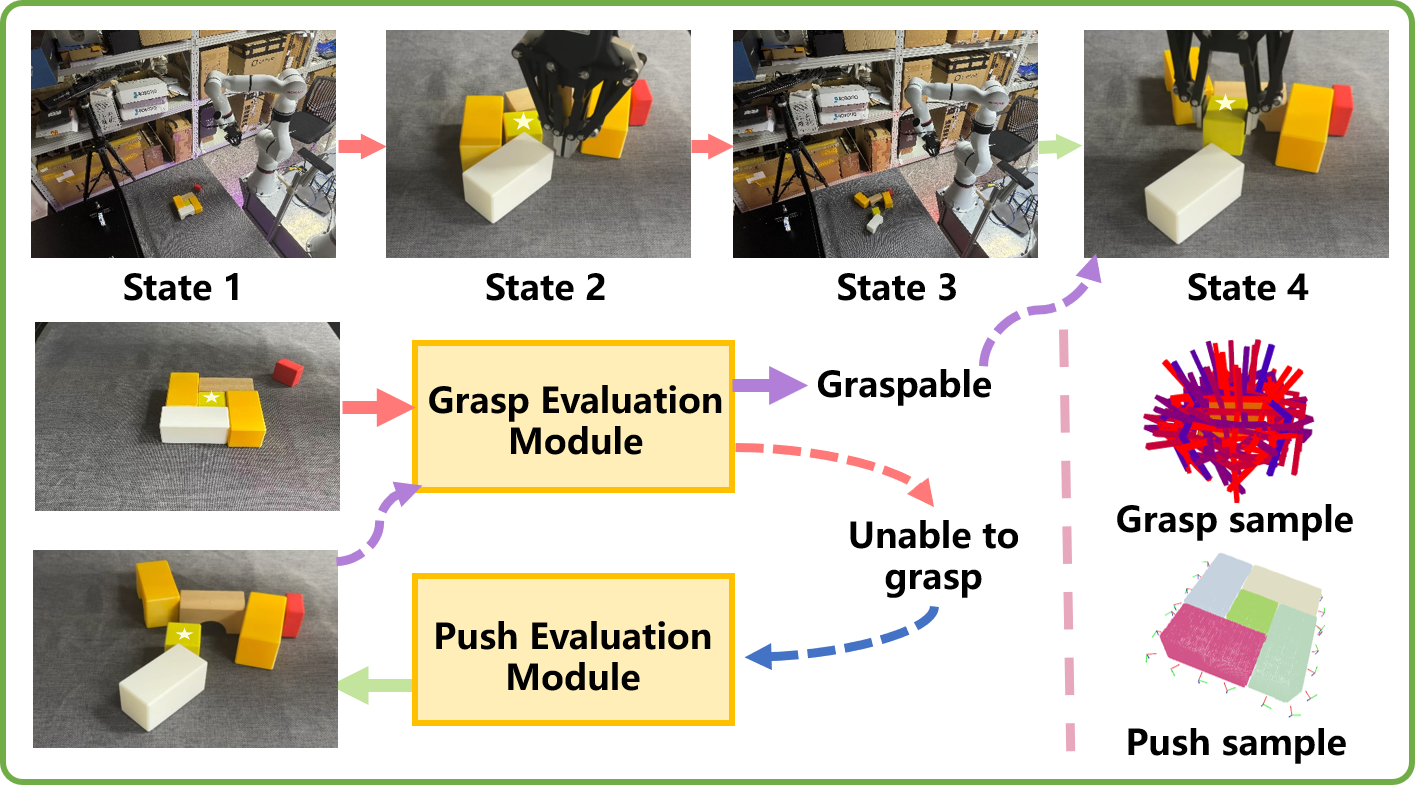} 
    \captionsetup{font=footnotesize} 
    \caption{This figure illustrates the GAPG workflow. The objective is to extract target objects with an asterisk label from dense scenes. Since the target object is not graspable in the initial state (State 1), we use pushing actions to adjust the object's spatial configuration to state (State 4). During this process, both grasping and pushing actions are derived from point cloud sampling. }
    \label{fig:overview}
\end{figure}

In cluttered environments, the available grasping space is often severely limited, and the grasping success rate heavily depends on the geometric match between the gripper and the surface of the target object, while also considering collisions with surrounding objects. Additionally, to improve push efficiency, it is crucial to account for the spatial geometric relationships between the target object and its environment. Ignoring such geometric information in 3D space can reduce grasp stability and overall push efficiency. However, there is a lack of research systematically leveraging 3D geometric information to address these issues.

To tackle these challenges, we propose a grasp evaluation module and a pushing evaluation module based on the PointNet++ architecture~\cite{b9}. The grasp evaluation module first converts the sampled grasp pose into the corresponding gripper point cloud as a reference for grasping. It then analyzes the geometric relationship between the point cloud of the gripper's closed space and the gripper's point cloud to assess grasp feasibility. The pushing evaluation module transforms the pushing pose into a push point, and under the guidance of the grasp evaluation module, it analyzes the effect of pushing on the spatial state of the target and surrounding objects to select the most effective pushing pose for execution. Our method fully exploits geometric perception in 3D space. Experimental results show that it performs exceptionally well in both simulation and real-world environments, demonstrating strong stability and generalization capability.The overall workflow is illustrated in Fig.~\ref{fig:overview}.

The main contributions of this work are:
\begin{itemize}
    \item We propose using GraspNet\cite{b8} to sample grasp poses and transform them into gripper point clouds. By analyzing the geometric relationship between the gripper point cloud and the point cloud inside the gripper’s closed space, we assess grasp feasibility, introducing a new safe and reliable grasping method for cluttered scenes.
    \item Using grasping as a guide, we transform the push pose into a push point, analyzing the geometric spatial relationship between the push point, the goal object, and surrounding objects. We implicitly learn how pushing actions affect the grasping space of the goal object, offering a new and efficient pushing method for cluttered scenes.
    \item We conduct extensive experiments in both simulation and real-world environments, validating the effectiveness and reliability of our approach. Notably, our method successfully transitions from simulation to real-world settings without requiring data fine-tuning and generalizes well to unseen objects.
\end{itemize}

\section{Related Work}

\subsection{Single Grasping}
In robotics, grasping tasks have a long history, covering key aspects such as grasp pose generation and scoring. Some methods use point clouds of objects as input, directly generating grasp poses and scoring them through end-to-end neural networks~\cite{b8,b11,b26,b33,b34}. Pointnetgpd\cite{b5} evaluates the feasibility of grasp poses based on an analysis of surface geometry. Yang et al.\cite{b12} proposes extracting point clouds around the target object to generate collision-free grasp configurations. Dexpoint\cite{b7} designs a reward function for gripper-object contact, using the PPO reinforcement learning algorithm to train grasping strategies. Additionally, methods ~\cite{b1, b2} use grasp success as a reward signal and apply Q-learning to learn grasping strategies.

Although models~\cite{b8,b25,b34} show good generalization and versatility in typical scenes, these general models often struggle to assess the graspability of target objects stably and accurately in cluttered and dense environments, where the grasping space is limited and the state is complex. On the other hand, methods~\cite{b2,b7} train grasping strategies using reinforcement learning for specific scenes. While effective in these particular environments, they lack stable generalization across diverse scenarios.

\subsection{Push-grasping Synery}
In many complex scenarios, a single grasp action is insufficient to complete the task, prompting researchers to explore collaborative grasping and pushing methods. These methods~\cite{b1,b2,b17,b18,b19} introduced pushing actions, using dual Q-networks to learn grasp and push selection strategies. Pushing helps break the non-graspable state of target objects in dense scenes, enabling clutter cleaning tasks without a specific target. Dipn\cite{b6} predicts the spatial changes caused by pushing when grasping is not feasible, then evaluates the graspability of the predicted state and selects the most beneficial pushing action for subsequent grasping. This model performs well in desktop cleaning tasks.

Compared to untargeted grasping, target-oriented grasping is more challenging. Xu et al.\cite{b2} built on Zeng et al.\cite{b1} by proposing a serial dual Q-network training mechanism for target-oriented grasping tasks. Mpgnet\cite{b10} further introduced a combined grasp-push auxiliary strategy, where, when the target object is obstructed or stacked, the robot first grasps the object on top and then adjusts the grasping space by pushing.

Most of these methods use heightmaps (2.5D) as input, achieving some success in specific scenes. However, by neglecting critical 3D geometric information in the grasping and pushing processes, the models' generalization ability is limited, and both grasp stability and pushing efficiency are compromised. Unlike the previous works, GarmentPile\cite{b4} uses 3D point clouds as input, employing both grasping and placement auxiliary actions to collaboratively adjust the grasping space. However, its auxiliary actions still rely on the supervision of an initial grasp feasibility judgment model. Due to the complexity of non-rigid objects, accurate grasping judgment is difficult to achieve with simple scene point clouds, leading to lower overall task efficiency.

Existing studies generally acknowledge the strong correlation between grasping and auxiliary actions, attempting to use grasp model evaluation signals to supervise auxiliary action training. However, these methods either ignore geometric structural information in 3D space or fail to effectively utilize such information. The proposed GAPG method uses point clouds as input, fully integrates spatial geometric perception, maintains stable grasping judgment, and achieves efficient pushing actions. It demonstrates strong generalization across multiple scenarios, effectively addressing the limitations of existing methods.

\section{Problem Formulation}
We define the target-oriented grasping task in cluttered scenes as a combination of grasping and pushing actions. In our experiments, we set up a planar workspace of size $50cm \times 45cm$ and place two depth cameras on the front and left sides. Given the scene point cloud $\mathbf{X} \in \mathbb{R}^{N \times 3}$ and the objects' masks, we sample two sets of action candidates:
\[
\mathcal{G} = \{(\mathbf{p}_i, \mathbf{R}_i)\}_{i=1}^{N_g}, 
\quad 
\mathcal{P} = \{(\mathbf{p}_j, \mathbf{R}_j)\}_{j=1}^{N_p},
\]
where $\mathbf{p} \in \mathbb{R}^3$ and $\mathbf{R} \in SO(3)$ denote 
the position vector and orientation (rotation) matrix of the gripper, respectively. 

The grasp evaluation module first assesses the feasibility of each grasp pose. 
If the predicted grasps score exceeds a pre-set threshold, the grasp pose with the highest score, 
$\mathbf{g}_{\text{best}} \in \mathcal{G}$, is executed. 
Otherwise, the push evaluation module
\begin{figure*}[t!]
    \centering
    \includegraphics[width=0.98\textwidth]{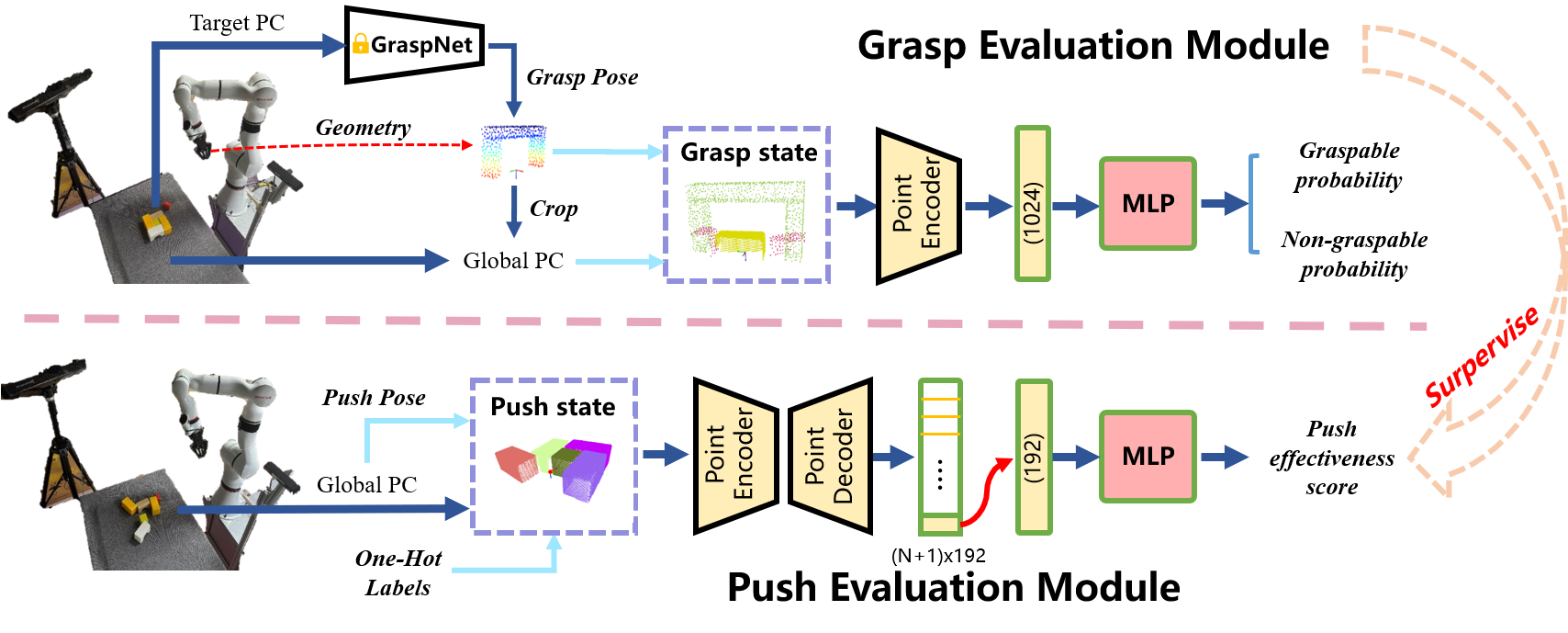} 
    \captionsetup{font=footnotesize} 
    \caption{Overview of our framework. \textbf{Grasp Evaluation Module}: The model takes as input the global point cloud and the target point cloud.Based on the grasp pose, a gripper point cloud is generated and concatenated with the point cloud within the gripper's closure space to form a grasp representation. This representation is then fed into a PointNet++ network to extract global geometric features, which are finally passed to an MLP for grasp feasibility analysis. \textbf{Push Evaluation Module}: The push pose is converted into a fixed push point, and the same spatial transformation is applied to the global point cloud. The transformed data is then concatenated with one-hot labels to distinguish the push point/target object from other objects. The synthesized push state is processed through PointNet++ to extract (N+1) × 192 features, from which the 1 × 192 feature corresponding to the push point is selected and fed into an MLP to score the push pose.
}
    \label{fig:model}
\end{figure*}
selects the push pose with the highest score, 
$\mathbf{p}_{\text{best}} \in \mathcal{P}$. The ultimate goal is to grasp the goal object without collisions.

\section{Method}

\subsection{Overview}
To address the challenges posed by complex and cluttered scenes, we design a  \textbf{Grasp Evaluation Module} , which performs geometric feature matching between the constructed gripper point cloud and the point cloud within the gripper’s closing space to achieve stable grasping. In addition, we further devise a \textbf{Push Evaluation Module} , which, under the supervision of the grasp evaluation module, implicitly learns the influence of push actions on the graspability state of the target object.

\subsection{Grasp Data Collection}
We employ the GraspNet\cite{b8} pre-trained model to generate 6D grasp poses. First, we obtain the point cloud of the target object through a mask and generate grasp poses exclusively for the target object. To maximize coverage of the target object with sampled actions, several geometric thresholds in GraspNet are relaxed. The strong generalization capability of GraspNet provides us with a diverse and plausible set of candidate grasp poses, which significantly improves training efficiency. In the simulation environment, we randomly load 10 objects and randomly select and execute actions from the candidate grasp set. If the target object is successfully grasped, the corresponding grasp pose label is set to 1; otherwise, it is set to 0. Following this strategy, we collected a dataset of 40,000 samples.

\subsection{Grasp Evaluation Module}
The architecture of our grasp evaluation module is illustrated in Fig.~\ref{fig:model}. Unlike PointNetgpd\cite{b5} which only considers
\begin{figure}[t!]
    \centering
    \includegraphics[width=0.48\textwidth]{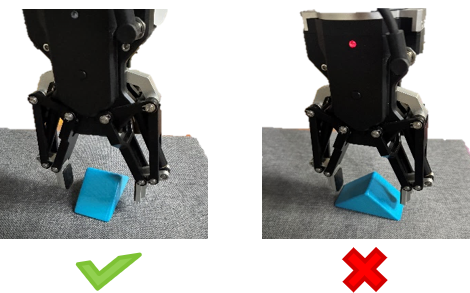} 
    \captionsetup{font=footnotesize} 
    \caption{Geometric matching between the gripper and the target object’s grasping surface: In the left figure, the gripper fingertips are parallel to the target object’s grasping surface, with a large contact area and uniform force distribution, resulting in a high geometric match, thus deemed graspable. In the right figure, when the gripper closes, the contact area with the target object is smaller, and the fingertips struggle to form a stable grasp on the surface, leading to a low geometric match, thus deemed non-graspable.}
    \label{fig:grasp_match}
\end{figure}
the point cloud within the gripper’s closing region while neglecting the gripper’s own geometric information, our method evaluates grasp feasibility by performing feature matching between the synthesized gripper point cloud and the point cloud within the closing region.  This design explicitly accounts for potential collisions between the gripper and surrounding objects, as well as the geometric alignment between the gripper and the target surface, As shown in Fig.~\ref{fig:grasp_match}.

To achieve a unified representation of grasp states, we transform the global point cloud into the grasp pose coordinate frame. Based on the fingertip dimensions of the real gripper, we then generate a virtual gripper point cloud centered at the origin of the gripper coordinate system. The point cloud of the closing region is subsequently extracted from the global point cloud using the gripper geometry. Finally, the synthesized gripper point cloud and the point cloud inside the closing region are concatenated into a single grasp state, ensuring that the gripper consistently serves as a reference basis for analyzing how variations within the closing region affect grasp outcomes.

To further distinguish between the gripper and the object being grasped, we augment the xyz point cloud representation with an additional channel, where a 0/1 label indicates whether a point belongs to the gripper or the closing region. The resulting grasp state is then fed into a PointNet++ network to extract a 1×1024 global geometric feature, which is passed through a three-layer MLP classifier to output the probabilities of “graspable” and “non-graspable.” The supervision labels are generated from actual grasp outcomes, and the module is trained using the CrossEntropyLoss\cite{b15}, defined as follows:
\begin{equation}
\mathcal{L}_{\text{grasp}}
= -\frac{1}{B}\sum_{i=1}^{B}
\left[
(1-\varepsilon)\,\log p_{i,y_i^*}
+ \frac{\varepsilon}{C}\sum_{c=1}^{C}\log p_{i,c}
\right]
\label{eq:grasp}
\end{equation}

In the task pipeline, the grasp evaluation module undertakes two key tasks: (1) determining whether the current state of the target object is graspable; and (2) providing supervisory labels for the execution of push actions.
\subsection{Push Data Collection}
Through 8000 data samples testing of the grasp evaluation module, we observed that when the graspability threshold is set to 0.8, the actual grasp success rate reaches 98.57\%. In the simulation environment, we randomly load scenes with 10 objects and filter out those that are directly graspable. The point clouds of all objects are obtained via segmentation masks, and push actions are sampled through spatial discretization. Specifically, we first project the object point cloud onto the $XY$ plane and extract the projection contour via morphological operations. Based on this contour, we dilate it outward by $0.016~\mathrm{m}$ to form a push sampling line, on which we sample push start locations with a fixed step size of $0.03~\mathrm{m}$ to obtain the planar coordinates $(x,y)$. The height $z$ is set to the vertical center of the object point cloud, i.e., the $z$-axis center of its oriented bounding box (OBB). The resulting $(x,y,z)$ constitutes the 3D push start point. For each start point, the pushing orientation is defined as follows: the $z$-axis points vertically downward; the $x$-axis lies in the horizontal plane and points from the start point projection toward the centroid of the projected object point cloud; and the $y$-axis is obtained as the cross product of the $z$- and $x$-axes. Finally, the orientation together with the start point is assembled into a $4 \times 4$ homogeneous transformation matrix representing the pushing pose.

Candidate push actions that may cause collisions are filtered using the global point cloud, and a random action is selected from the remaining candidates for execution. A push action is considered effective (label = 1) only if it alters the graspable space of the target object and the grasp evaluation module subsequently assigns a graspability score greater than 0.8; otherwise, it is labeled 0. Following this sampling strategy, we generated 40,000 data samples to train the push evaluation module.
\subsection{Push Evaluation Module}
As illustrated in Fig.~\ref{fig:model}, the architecture of our push evaluation module is designed to capture the influence of push actions on both the target object and surrounding objects. We abstract a push action as a point with pose information $\mathcal{P} = (x, y, z, R)$. Within the workspace, a fixed reference pose $\mathcal{P'} = (x',y',z,R')$ is defined to represent the planar position and orientation of the push. Here, we set $x' = 0.5$, $y' = 0$, and set $R'$ to be aligned with the robot base frame. The push pose is then transformed into this fixed reference pose ($T \cdot \mathcal{P} \rightarrow \mathcal{P'}$). In particular, the $z$-coordinate is kept unchanged during this transformation. Subsequently, the same transformation is applied to the point cloud ($T \cdot points \rightarrow points'$). This modification ensures that the push action always originates from the fixed reference
\begin{figure}[t!]
    \centering
    \includegraphics[width=0.48\textwidth]{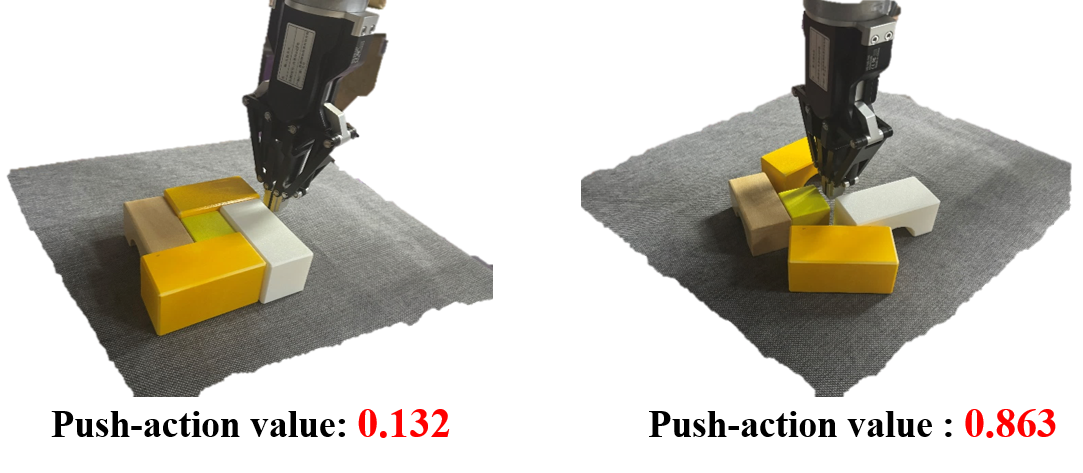} 
    \captionsetup{font=footnotesize} 
    \caption{Push action value score: In the left figure, due to the highly dense state, it is difficult to make the target object graspable through the current push, resulting in a low score . In the right figure, in a moderately dense state, the current push can enable the target object to reach a graspable state in the next step, thus resulting in a higher score.}
    \label{fig:push_learning}
\end{figure}
point and moves a constant distance of 0.125m along the positive $x$-axis.  

To better analyze the spatial effects of push actions, we assign one-hot labels to different point types: $[1,0,0]$ for the push point, $[0,1,0]$ for the target object, and $[0,0,1]$ for other objects. These labels are concatenated with the original point cloud, resulting in a push state of size $(N+1) \times 6$. The push state is then fed into the PointNet++ network to extract $(N+1) \times 192$ features. We retain only the $1 \times 192$ feature corresponding to the push point, which implicitly encodes the impact of the push action on both the target object and surrounding objects. As shown in Fig.~\ref{fig:push_learning}, the extracted feature is subsequently processed by a three-layer MLP to classify the push point, producing the push effectiveness probability.  

The supervision labels are derived from the actual results of push executions, which are determined by the predicted grasp feasibility scores of the grasp module, and the module is trained using Binary Cross-Entropy Loss (BCELoss), defined as follows:
\begin{equation}
\mathcal{L}_{push} = - \frac{1}{N} \sum_{i=1}^N y_i \log \hat{y}_i + (1-y_i)\log (1-\hat{y}_i)
\end{equation}
 
In cases where the target object is not directly graspable, we select the push action with the highest push effectiveness probability from the sampled candidates and execute it to alter the graspable state of the target object.
\begin{figure}[!t]
    \centering
    \includegraphics[width=0.48\textwidth]{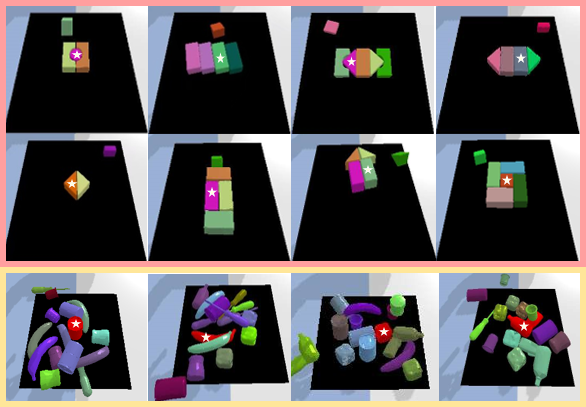} 
    \captionsetup{font=footnotesize} 
    \caption{In the PyBullet simulation environment, eight artificially created challenge scenes and four scenes loaded randomly with 15 unseen objects. The objects marked with an asterisk in these scenes are the target objects. }
    \label{fig:challenge}
\end{figure}
\subsection{Training and Inference Details}
Data collection was conducted on an Nvidia RTX A6000, where 40,000 samples were collected for both the grasp module and the push module, module training was carried out on an NVIDIA GeForce 4090.

For the grasp evaluation module, the training batch size was set to 256. The initial learning rate was $5 \times 10^{-4}$, with a total of 85 training epochs. In the 40th, 55th, and 80th epochs, the learning rate was reduced to 0.1 of its original value. AdamW\cite{b13} optimizer was used during training. For the push evaluation module, the training batch size was set to 128 and the validation batch size to 64. The initial learning rate was $8 \times 10^{-4}$, with 100 training epochs. Whenever the validation loss did not decrease, the learning rate was updated with a decay factor of 0.95. The Adam\cite{b14} optimizer was used during the training. 

During inference with the grasp evaluation module, all sampled candidate actions are aggregated into a grasp state set and input to the module as a batch. The push evaluation module follows the same procedure.

\section{Experiments}
We design a series of experiments to verify the following:  
(1) our grasp evaluation module can accurately and reliably 
determine whether a target object is graspable;  
(2) our push evaluation module can efficiently and robustly optimize the graspable state of the target object;  
(3) our model exhibits strong generalization ability, performing well both in sim-to-real transfer and on unseen objects, without requiring any additional data fine-tuning.

\subsection{Evaluation Metrics}
We designed a series of scenarios to validate our model, the simulation environments are partially illustrated in Fig.~\ref{fig:challenge}, including:
\begin{figure}[t!]
    \centering
    \begin{subfigure}{1.0\linewidth}  
        \centering
        \includegraphics[width=0.8\linewidth]{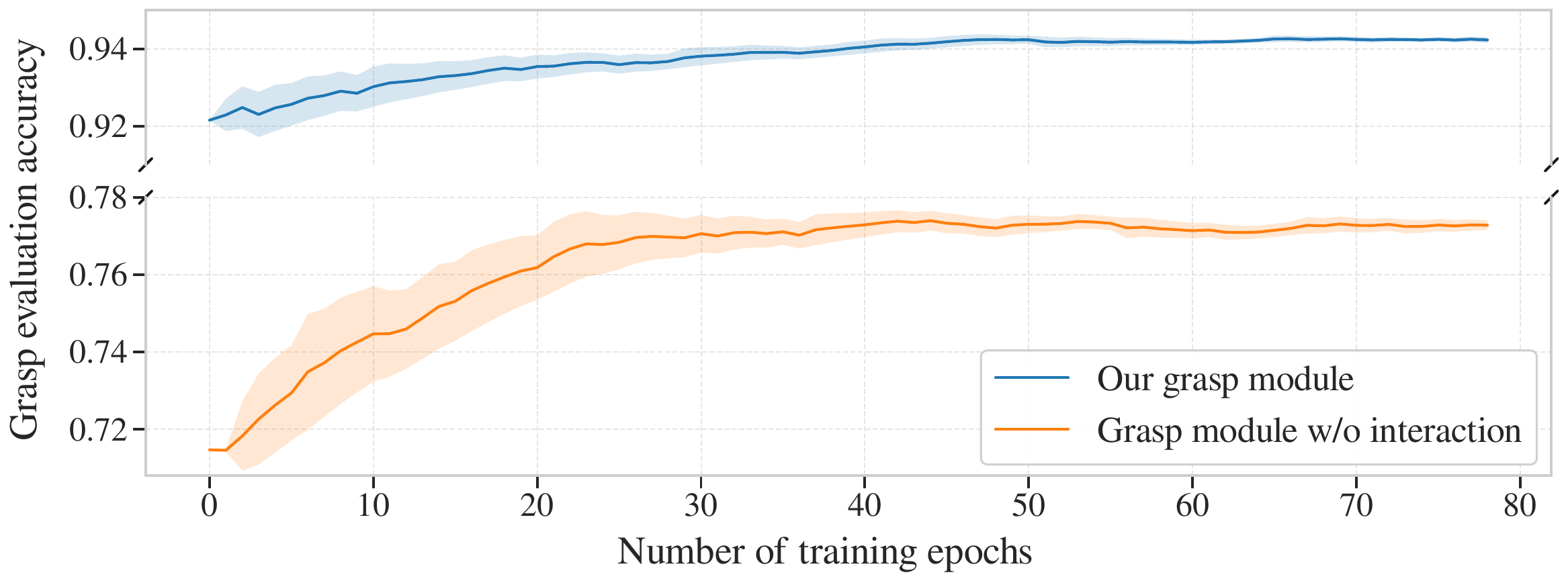} 
        \caption{Grasp module training curve}
        \label{fig:grasp_train}
    \end{subfigure}
    
    \par\bigskip 
    
    \begin{subfigure}{1.0\linewidth}  
        \centering
        \includegraphics[width=0.8\linewidth]{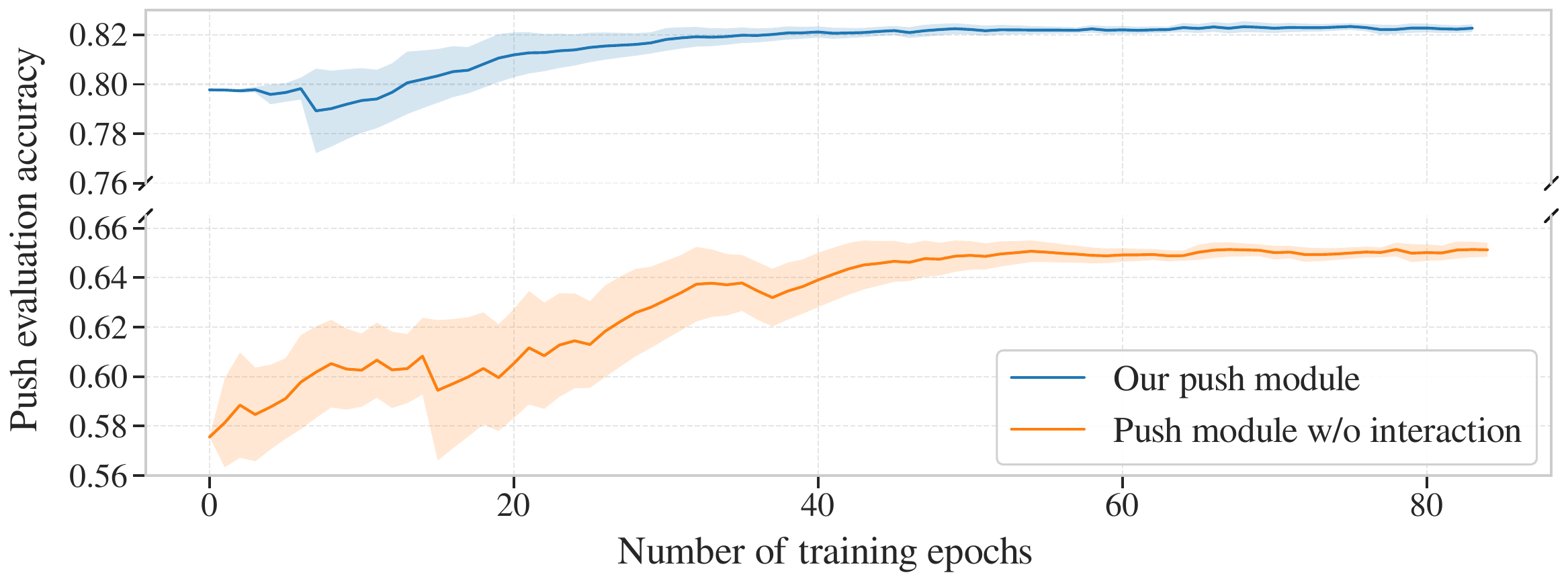} 
        \caption{Push module training curve}
        \label{fig:push_train}
    \end{subfigure}
    \captionsetup{font=footnotesize} 
    \caption{Module training and ablation experiments: The module converged after 80 training epochs. Visually, our system outperforms the ablation methods by a significant margin, both in grasp evaluation accuracy and push evaluation accuracy.}
    \label{fig:train_curves}
\end{figure}
\begin{enumerate}[]
    \item 15 lightly cluttered objects,
    \item 30 heavily cluttered objects,
    \item 15 scenes with unseen objects,
    \item challenge scenes.
\end{enumerate}

We evaluate our model using three types of metrics:  
\begin{itemize}
    \item \textbf{Completion Rate}: If the robot successfully grasps the target object without collision within ten combined push and grasp attempts, the trial is counted as successful.  
    \item \textbf{Grasp Success Rate}: Given $n$ test trials in a scene, when the grasp evaluation module predicts the target object as graspable, we record whether the executed grasp succeeds without collision. 
    \item \textbf{Motion Number}: The number of executed grasp and push actions required for task completion.  
\end{itemize}

\subsection{Simulation Experiments}
Xu et al.\cite{b2} is the most closely related to our task, and therefore we adopt it as the baseline for comparison. To ensure fairness, we construct identical scenarios on the PyBullet\cite{b24} simulation platform for both training and inference. Due to differences in training procedures, our comparison mainly focuses on inference results.  

For training the grasp evaluation module and the push evaluation module, we randomly load 10 objects in the scene, acquire full point clouds from multiple views, and add random Gaussian noise. The point clouds are then downsampled to 345 points for the grasp evaluation module and 1024 points for the push evaluation module. After 80 training epochs, the accuracy of the grasp evaluation module stabilizes above 94\%, while the push evaluation module stabilizes above 82\%, as shown in Fig.~\ref{fig:train_curves}.  

\begin{figure*}[htbp]
    \centering
    \includegraphics[width=1\textwidth]{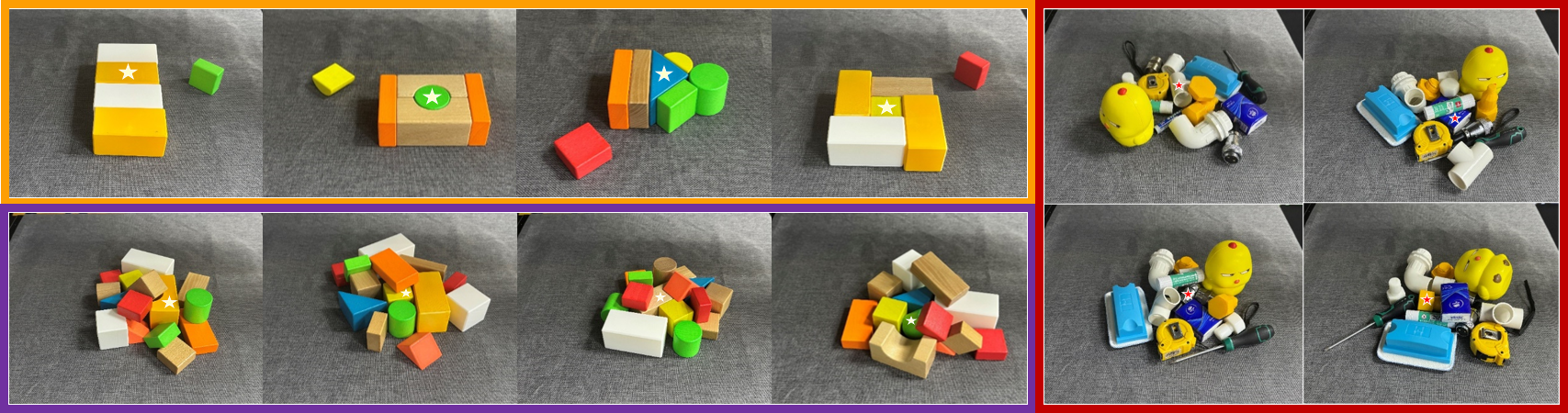} 
    \captionsetup{font=footnotesize} 
    \caption{Three types of scenes were arranged in the real-world setting, with the target objects marked by an asterisk. Orange frame: challenge scenes manually constructed; Purple frame: cluttered and dense scenes with 15 objects manually arranged; Red frame: 15 objects not seen in the training dataset and with geometric shapes significantly different from those in the training objects.}
    \label{fig:real_world_setup}
\end{figure*}

\begin{table*}[htbp]
\centering
\caption{Comparison of different methods in simulation}
\label{tab:sim}
\begin{tabular}{lccc|ccc|ccc}
\toprule
\multirow{2}{*}{Method} & \multicolumn{3}{c}{Task success rate} & \multicolumn{3}{c}{Grasp success rate} & \multicolumn{3}{c}{Average motion number} \\
\cmidrule(lr){2-4} \cmidrule(lr){5-7} \cmidrule(lr){8-10}
 & 15 objects & 30 objects & challenge & 15 objects & 30 objects & challenge & 15 objects & 30 objects & challenge \\
\midrule
GAPG-Single Grasping           & 43.3\% & 13.3\% & / & 100\% & 100\% & / & / & / & / \\
Xu et al.\cite{b2}-Single Grasping   & 26\% & 12\% & / & 85\% & 79.3\% & / & / & / & / \\
Efficient Push Grasping \cite{b2} & 81\% & 76\% & 86\% & 84\% & 80.1\% & 87\% & 3.93 & 4.93 & 2.93 \\
GAPG w/o gripper pc  & 76.7\% & 60\% & 90\% & 76.7\% & 64.3\% & 90\% & 2.90 & 5.00 & 2.60 \\
GAPG                  & 100\% & 100\% & 100\% & 100\% & 96.7\% & 100\% & 1.63 & 2.80 & 2.33 \\
\bottomrule
\end{tabular}
\end{table*}

\begin{figure*}[htbp]
    \centering

    \begin{subfigure}[t]{0.48\linewidth}
        \centering
        \includegraphics[width=\linewidth]{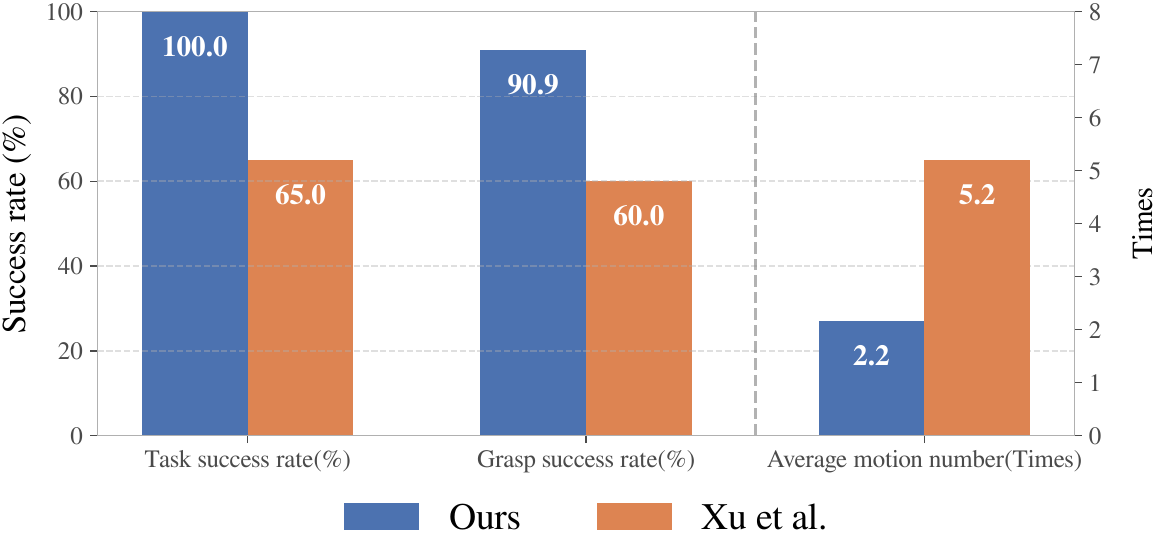}
        \caption{sim}
        \label{fig:sim}
    \end{subfigure}
    \hfill
    \begin{subfigure}[t]{0.48\linewidth}
        \centering
        \includegraphics[width=\linewidth]{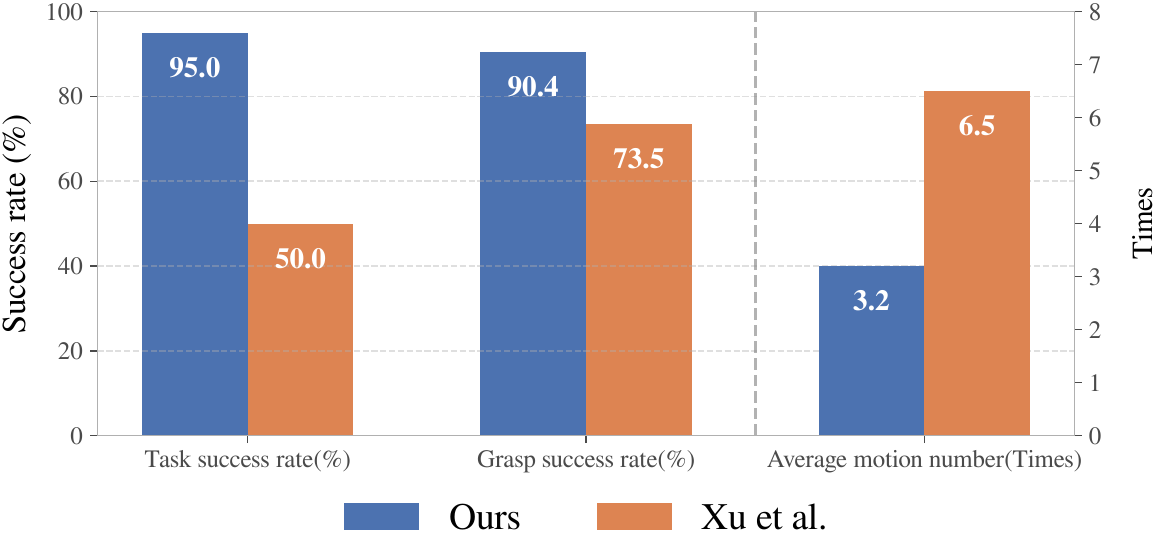}
        \caption{real world}
        \label{fig:real_world}
    \end{subfigure}
    \captionsetup{font=footnotesize} 
    \caption{Results of generalization experiments on unseen objects. Figure~(a) shows the experimental results in a simulated environment with a scene containing 15 unseen objects; Figure~(b) presents the experimental results in a real-world scenario with a physically arranged scene of 15 unseen objects.}
    \label{fig:unseen}
\end{figure*}
During inference, we set identical random seeds to ensure consistent environment initialization, and evaluate the performance of both \textit{single grasp} and \textit{grasp with push} strategies. It should be noted that, unlike the training stage where only 10 objects are loaded, we directly generalize the models to more complex and cluttered scenes with 15 and 30 objects, and set the graspability threshold of the grasp evaluation module to 0.8. Each scenario type (15 objects, 30 objects, and manually designed challenge scenes) is tested 30 times, and the average experimental results are summarized in Table~\ref{tab:sim}.  

The results demonstrate that our model outperforms the baseline across all evaluation metrics. In particular, in both the 15-object and 30-object scenes, our model achieves a grasp success rate exceeding 95\%, which is approximately 20\% higher than the baseline. This indicates that our grasp evaluation module possesses greater stability and generalization capability. However, relying on grasping alone, both our method and the baseline fail to effectively accomplish the tasks.  

Furthermore, in the 30-object random scenes, our model requires an average motion number of only 2.8, which improves task efficiency by about 75\% over the baseline (measured as $\frac{\text{baseline} - \text{GAPG}}{\text{GAPG}}$). This highlights that our push evaluation module can efficiently transform non-graspable states into graspable ones. With the combination of efficient pushing and stable grasping, our model achieves a 100\% task success rate across all scenarios, while also attaining the highest task efficiency. To further verify the generalization capability of our model, we randomly loaded 15 unseen objects in the simulation environment for testing. The experimental results are shown in Fig.~\ref{fig:unseen}. As can be seen from the results, all of our metrics outperform the baseline, and the task completion rate reaches 100\%. These findings suggest that our proposed model, by analyzing the three-dimensional geometric relationships involved in grasping and pushing, is better suited to complex and cluttered scenes, exhibiting superior stability.
\subsection{Ablation Study}
We design two variants: (1) removing the gripper point cloud in the grasp evaluation module (without gripper pc in grasp evaluation module), i.e., retaining only the local point cloud within the grasping region, similar to the method in \cite{b5}, in order to investigate the impact of the gripper point cloud on the performance of the grasp evaluation module; (2) removing the gripper point cloud in the GAPG framework (without gripper pc in GAPG), i.e., using the grasp evaluation module without the gripper point cloud as supervision for training the push evaluation module, to study how adjustments in the grasp evaluation module affect the push evaluation module.  

The training process uses the same dataset, with the only difference being the absence of the gripper point cloud. The training results are shown in Fig.~\ref{fig:train_curves}. It can be clearly observed that without geometric gripper information, the training of the grasp evaluation module becomes more oscillatory, and its accuracy decreases by 22.23\%. Furthermore, the reduced accuracy of the grasp evaluation module leads to a 20.7\% drop in the accuracy of the push evaluation module.  

To further quantify the performance gap, we randomly load 15 or 30 objects in the simulation environment, as well as manually designed challenge scenarios, and conduct 30 inference trials for each case. The average results are reported in Table~\ref{tab:sim}. In random scenes, the grasp success rate of GAPG without gripper pc drops by more than 20\%. This is because the absence of gripper point cloud information prevents precise judgment of grasp feasibility and causes more frequent collisions during grasping. The push efficiency also decreases by 75\% (calculated as  $\frac{\text{GAPG without gripper pc} - \text{GAPG}}{\text{GAPG}}$). The main reason is the lack of reliable grasp supervision, even when a push action transforms the target object into a fully graspable state, the misjudgment of the grasp evaluation module can still result in grasp failure, leading to more ineffective pushes.  

In summary, by introducing geometric matching between the gripper point cloud and the local grasping region, we ensure the stability and accuracy of the grasp evaluation module, while providing effective supervision for the push evaluation module, thereby enabling stable and efficient task completion.

\subsection{Real-World Experiments}
To validate the strong generalization ability of our model, we design three types of test scenarios in real-world environments: four randomly arranged scenes with 15 seen objects, four manually designed challenge scenes, and four randomly arranged scenes with 15 unseen objects, as illustrated in Fig.~\ref{fig:real_world_setup}.  

The experimental hardware setup is as follows: the robot arm is a Rokae xMate ER7Pro, and the gripper is a DH AG-105-145. It should be noted that, unlike in the simulation environment where the full point cloud is available, in real-world inference we only use two cameras: a Photoneo PhoXi 3D Scanner and a ZED 2i camera. We extract the target object point cloud using the mask generated by SAM\cite{b32}.The entire system is integrated through ROS, with inverse kinematics solved by MoveIt!\cite{b31}.  

For each scenario, we conduct five trials. If a collision occurs during grasping, the task is immediately considered
\begin{table}[t!]
\centering
\caption{Comparison of different methods in real-world}
\label{tab:real}
\resizebox{0.98\linewidth}{!}{ 
\begin{tabular}{lcc}
\toprule
Method & Efficient Push-Grasping & GAPG \\
\midrule
Task success rate (15 objects)  & 65\% & 95\% \\
Task success rate (challenge)   & 70\% & 100\% \\
Grasp success rate (15 objects) & 81.3\% & 100\% \\
Grasp success rate (challenge)  & 82.4\% & 100\% \\
Average motion number (15 objects) & 5.75 & 2.75 \\
Average motion number (challenge) & 4.60  & 2.50 \\
\bottomrule
\end{tabular}
}
\end{table}
\noindent a failure. The results are summarized in Table~\ref{tab:real}. The experimental results demonstrate that our model significantly outperforms the baseline across all metrics. In the random scenes, both the task completion rate and the grasp success rate reach 95\%. In the manually designed challenge scenes, all metrics achieve 100\%, fully showcasing the stability and generalization capability of our model. To further demonstrate our model's generalization capability in real-world scenarios, 15 unseen objects with significantly different geometric shapes from the training objects were selected for testing. As shown in Fig.~\ref{fig:unseen}, our model achieves a 95\% task completion rate and a 90\% grasp success rate even in highly cluttered environments with completely novel objects. 

We further observe that most failures are caused by calibration errors of the cameras, which prevent the robotic arm from accurately reaching the intended positions. In relatively dense scenes, such errors are more likely to result in collisions, ultimately leading to task failure.  

In summary, our model is able to maintain stable grasping performance and efficient pushing capability when transferred to real-world environments without fine-tuning. Compared with the baseline, this further highlights the importance of effectively leveraging three-dimensional spatial geometric information for tasks in complex and cluttered environments.

\section*{Conclusion and Future Work}
In this work, we propose the GAPG model, which relies entirely on point clouds as input. We transform the evaluation of grasping and pushing actions into the analysis of geometric relationships between points in the cloud, providing a novel approach for operations in cluttered environments. Extensive experimental validation demonstrates that our grasp evaluation module exhibits high generalization and strong stability. Based on this, we provide high-quality supervision for the training of the downstream pushing evaluation module, ensuring the efficiency of pushing actions. Experimental results show that our model significantly outperforms methods based on heightmaps, which is likely due to the effective use of 3D spatial geometric information, making it more suitable for tasks in dense environments. In the future, we plan to develop two auxiliary actions, grasp and push, to address more complex scenarios.

\bibliographystyle{ieeetr}   
\bibliography{References} 

\end{document}